\documentclass[a4paper,12pt]{article}
\usepackage[utf8]{inputenc}
\usepackage[T1]{fontenc} 
\usepackage{lmodern}     
\usepackage{graphicx}
\usepackage{amsmath}
\usepackage{geometry}
\usepackage{setspace}
\usepackage{titlesec}
\usepackage{authblk}
\usepackage{adjustbox}
\usepackage{lipsum}
\usepackage[backend=biber,style=numeric,maxcitenames=2,maxbibnames=6,giveninits,uniquename=init]{biblatex}
\usepackage{subcaption}
\usepackage{booktabs}
\usepackage{microtype} 
\usepackage[breaklinks=true]{hyperref}
\usepackage{xurl}

\newcommand{\affilA}{Biological and Agricultural Engineering, Kansas State University, Manhattan, Kansas 66506, USA}
\newcommand{\affilB}{Department of Agricultural Machinery and Technologies Engineering, Agriculture Faculty, Selçuk University, Konya, 42060, Turkey (TR)}

\title{\bfseries\Large FruitProm: Probabilistic Maturity Estimation and Detection of Fruits and Vegetables} 

\author[1]{Sidharth Rai \thanks{Equally Contributing Author}}         
\author[1]{Rahul Harsha Cheppally $^*$}
\author[1]{Benjamin Vail}
\author[2]{Keziban Yalçın Dokumacı}
\author[1]{Ajay Sharda} 

\affil[1]{\affilA} 
\affil[2]{\affilB}
\affil[ ]{\small Emails: \url{sidharth@ksu.edu} (SR), \url{r4hul@ksu.edu} (RHC), \url{benv86@ksu.edu} (BV), \url{kezibanyalcin@selcuk.edu.tr} (KYD), \url{asharda@ksu.edu} (AS)}

\date{} 

\begin{document}

\maketitle

\begin{minipage}{\textwidth}
\section*{Abstract}
\noindent
Maturity estimation of fruits and vegetables is a critical task for agricultural automation, directly impacting yield prediction and robotic harvesting. Current deep learning approaches predominantly treat maturity as a discrete classification problem (e.g., unripe, ripe, overripe). This rigid formulation, however, fundamentally conflicts with the continuous nature of the biological ripening process, leading to information loss and ambiguous class boundaries. In this paper, we challenge this paradigm by reframing maturity estimation as a continuous, probabilistic learning task. We propose a novel architectural modification to the state-of-the-art, real-time object detector, RT-DETRv2, by introducing a dedicated probabilistic head. This head enables the model to predict a continuous distribution over the maturity spectrum for each detected object, simultaneously learning the mean maturity state and its associated uncertainty. This uncertainty measure is crucial for downstream decision-making in robotics, providing a confidence score for tasks like selective harvesting. Our model not only provides a far richer and more biologically plausible representation of plant maturity but also maintains exceptional detection performance, achieving a mean Average Precision (mAP) of 85.6\% on a challenging, large-scale fruit dataset. We demonstrate through extensive experiments that our probabilistic approach offers more granular and accurate maturity assessments than its classification-based counterparts, paving the way for more intelligent, uncertainty-aware automated systems in modern agriculture

\medskip 
\noindent\textbf{Keywords:} artificial intelligence, computer vision, object detection, YOLO26, RTDETR, RTDETV2, DIEM, Agriculture, Fruit, Vegetable, Maturity, Estimation, Detection, Probabilistic, Uncertainty, Quantification
\end{minipage}

\section{Introduction}

The capabilities of modern automated systems have been fundamentally redefined by the paradigm shift in artificial intelligence (AI), catalyzed by transformative advancements in machine learning (ML) and computer vision (CV)\cite{lindrothAppliedArtificialIntelligence2024}. These technologies have enabled the development of sophisticated systems capable of extracting and interpreting meaningful, real-time information from visual data sources, thereby automating tasks that were historically reliant on subjective human visual assessment. \cite{caoReviewComputerVision2025}
This technological revolution has catalyzed significant improvements across domain, including autonomous driving \cite{cheppallyRowDetrEndtoEndRow2025,yurtseverSurveyAutonomousDriving2020, masmoudiObjectDetectionLearning2019,hnewaObjectDetectionRainy2021}, healthcare diagnostics \cite{elakkiyaCervicalCancerDiagnostics2022,mercaldoObjectDetectionBrain2023,estevaDermatologistlevelClassificationSkin2017}, intelligent surveillance \cite{mishraStudyVideoSurveillance2016} and agriculture \cite{kamilarisDeepLearningAgriculture2018,raiEnhancingSeedingEfficiency2025, dalalComputerVisionBased2025, vailAxisAligned3DStalk2025}.

At the heart of this revolution lies the task of object detection, which involves identifying and localizing objects within images or video frames \cite{zhuReviewVideoObject2020}. The evolution of methodologies in this domain has been remarkably swift, progressing through several distinct architectural paradigms. Early approaches, rooted in traditional computer vision, relied on hand-crafted, feature-based methods such as the Histogram of Oriented Gradients (HOG),Viola-Jones cascades \cite{violaRapidObjectDetection2001} and sparse coding models \cite{renHistogramsSparseCodes2013,}. While these methods established important foundational principles, they have been largely rendered obsolete by the deep learning era, which introduced end-to-end learning frameworks that eliminated the need for manual feature engineering.

The modern  era of object detection was ushered in by \textbf{Convolutional Neural Networks (CNNs)}\cite{osheaIntroductionConvolutionalNeural2015, simonyanVeryDeepConvolutional2015,szegedyGoingDeeperConvolutions2014,heDeepResidualLearning2015,krizhevskyImageNetClassificationDeep2012}, whose architecture is inherently suited to learning spatial hierarchies from images \cite{guRecentAdvancesConvolutional2018}. Seminal architectures such as the Region-based CNN (R-CNN) family \cite{girshickRichFeatureHierarchies2014a,girshickFastRCNN2015,renFasterRCNNRealTime2016,heMaskRCNN2018} established the efficacy of two-stage  pipelines, achieving unprecedented accuracy benchmarks through region proposal mechanisms followed by classification and refinement \cite{liuDeepLearningGeneric2020}. 

 In parallel, the \textbf{You Only Look Once (YOLO)} series pioneered single-stage detection model architectures, delivering real-time performance that enabled widespread deployment in latency-critical applications  \cite{girshickRichFeatureHierarchies2014, redmonYouOnlyLook2016, wangYOLOv9LearningWhat2024,jocherUltralyticsYOLO2023,reisRealTimeFlyingObject2024,wangYOLOv7TrainableBagofFreebies2023a,liYOLOv6SingleStageObject2022,jocherYOLOv5Ultralytics2020}.
 Building upon these CNN foundations, \textbf{Transformer-based architectures} adapted the self-attention mechanism from natural language processing to the visual domain \cite{vaswaniAttentionAllYou2023, dosovitskiyImageWorth16x162021a,khansalmanTransformersVisionSurvey2022}. The Detection Transformer (DETR) \cite{carionEndtoEndObjectDetection2020} introduced an elegant end-to-end formulation that eliminated hand-crafted components such as anchor generation and non-maximum suppression, enabling more holistic scene understanding through global attention mechanisms. Subsequent refinements have addressed computational efficiency and detection performance, including Deformable DETR \cite{zhuDeformableDETRDeformable2021}, which introduced deformable attention modules for faster convergence, LW-DETR (Lightweight DETR) \cite{chenLWDETRTransformerReplacement2024}, optimized for resource-constrained deployments, Salience DETR \cite{houSalienceDETREnhancing2024a}, which incorporates saliency-guided attention mechanisms for improved small object detection, and the Real-Time Detection Transformer (RT-DETR) \cite{zhaoDETRsBeatYOLOs2024}, which achieves YOLO-competitive inference speeds while maintaining transformer advantages. The most recent iteration, RT-DETRv2 \cite{lvRTDETRv2ImprovedBaseline2024}, further advances real-time performance through architectural optimizations and enhanced training strategies. Beyond pure detection architectures, transformer-based approaches have enabled powerful co-design frameworks such as Co-DETR (Collaborative Hybrid Assignments Training) \cite{zongDETRsCollaborativeHybrid2023}, which combines multiple assignment strategies for improved training efficiency and detection accuracy.

The evolution of Vision-Language Models has progressed rapidly from foundational contrastive learning to sophisticated multimodal reasoning systems. CLIP \cite{radfordLearningTransferableVisual2021a} established the paradigm of vision-language alignment through contrastive pre-training on 400 million image-text pairs, enabling zero-shot transfer capabilities. This foundation enabled subsequent architectural innovations including BLIP \cite{liBootstrappingLanguageImagePretraining2022}, which introduced captioner-filter bootstrapping for handling noisy web data, and Flamingo \cite{alayracFlamingoVisualLanguage2022}, which pioneered few-shot learning through perceiver resamplers and gated cross-attention. The BLIP-2 architecture \cite{liBlip2BootstrappingLanguageImage2023} achieved efficient frozen model alignment via its Q-Former, bridging vision encoders and large language models with 54x fewer trainable parameters. Grounded understanding advanced through GLIP \cite{liGroundedLanguageImagePretraining2022}, which unified object detection with phrase grounding, followed by Grounding DINO \cite{liuGroundingDINOMarrying2024}, which combined DINO's detector architecture with grounded pre-training for open-vocabulary detection, and YOLO-World \cite{chengYOLOWorldRealTimeOpenVocabulary2024}, which brought real-time open-vocabulary detection capabilities. LLaVA \cite{liuVisualInstructionTuning2023} democratized VLM research as the first easily reproducible open-source model using visual instruction tuning, while GPT-4V \cite{openaiGPT4VisionSystem2023} set new benchmarks for commercial multimodal systems. Microsoft's Florence-2 \cite{xiaoFlorence2AdvancingUnified2023} presented a unified, prompt-based architecture for diverse vision-language tasks trained on 5.4 billion annotations. The field has since evolved into two distinct paradigms: Large Multimodal Models (LMMs) and Vision-Language-Action (VLA) models. Current proprietary LMMs include Gemini 2.5 Pro \cite{googleGemini25Pro2025}, which leads on reasoning benchmarks, GPT-5 \cite{openaiGPT52025}, featuring unified reasoning and response modes, and Claude Sonnet 4.5 \cite{anthropicClaudeSonnet452025}, excelling at agentic computer use. The open-source ecosystem has achieved comparable performance through models like Qwen3-VL-235B \cite{baiQwen3VL2025}, InternVL3.5-241B \cite{chenInternVL352025}, and Llama 4 Scout and Maverick \cite{metaLlama4Herd2025}, which employ mixture-of-experts architectures with native multimodal pre-training. The VLA paradigm, which extends vision-language understanding to embodied robot control, emerged with RT-2 \cite{broylesRT2VisionLanguageActionModels2023}, the first model to co-fine-tune large vision-language models on robotics data. Recent VLA advances include $\pi0$ \cite{blackPi0GeneralistPolicy2024}, demonstrating dexterous manipulation through flow matching, Helix \cite{figureaiHelixVisionLanguageAction2025}, the first full-body humanoid VLA, and Groot N1 \cite{nvidiaGrootN12025,yangQwen3TechnicalReport2025}, optimized for humanoid platforms with 10ms latency reactive control.

However, this progression towards increasingly sophisticated and powerful models has introduced a critical trade-off: computational complexity and inferential latency.\cite{tanEfficientNetV2SmallerModels2021} The multi-stage pipelines characteristic of R-CNN variants and the substantial computational demands of large-scale VLMs render them impractical for applications requiring immediate response\cite{huangSpeedAccuracyTradeOffs2017}. In contrast, the architectural efficiency of optimized single-stage detectors such as the YOLO series and streamlined Transformers like RT-DETR \cite{zhaoDETRsBeatYOLOs2024} has maintained their dominance in real-time systems. For deployment on resource-constrained edge devices or time-sensitive in robotic systems, where low latency is not just a preference but a necessity, these efficient architectures remain the most viable solution.

Among the many domains being transformed by these real-time vision technologies, few carry the  societal significance and global urgency of agriculture. Confronted with a projected global population approaching ten billion by 2050, the agricultural sector faces the formidable challenge of substantially increasing food production while contending with escalating resource scarcity and the unpredictable effects of climate change \cite{unWorldPopulationProspects2022}. In response to these pressures, the paradigm of Precision Agriculture has emerged as a transformative approach, leveraging data-driven technologies to optimize resource inputs, maximize yields, minimize environmental impact, and enhance overall sustainability\cite{BenefitsEvolutionPrecision}. The deployment of intelligent systems for tasks such as automated pest detection \cite{lippiYOLOBasedPestDetection2021,albaneseAutomatedPestDetection2021}, robotic harvesting and survey \cite{cheppallyRowDetrEndtoEndRow2025,DesignFieldEvaluation}, yield prediction \cite{vanklompenburgCropYieldPrediction2020}, and site-specific nutrient \cite{sultanResourceManagementAgriculture2023} management is no longer a futuristic concept but a present-day necessity for ensuring global food security \cite{kamilarisReviewPracticeBig2017}.

A common thread uniting these advanced agricultural applications is the fundamental requirement for timely, accurate, and objective data regarding crop health and development status. However, the primary method for gathering this data faces a critical bottleneck. The effective management of agricultural systems is critically dependent on an accurate understanding of crop phenology, or maturity stage \cite{meierBBCHSystemCoding2009}. Historically, this assessment has been performed through manual, in-field observation. This long-standing practice is widely cited as being laborious, subjective, and error-prone \cite{sadeghi-tehranAutomatedMethodDetermine2017}. This subjectivity is not a minor concern but a "real liability" in commercial and research settings, as "different observers may likely perceive the growth stage of the same plot differently," introducing significant human error into critical datasets \cite{HowMASSeeds, sadeghi-tehranAutomatedMethodDetermine2017}. The challenge is compounded by the ambiguous nature of biological development, where the transition between successive stages can be difficult to differentiate, directly compromising tasks like yield prediction \cite{puranbridgemohanApplicationBBCHScale2016}.

While computer vision and deep learning offers a technologically compelling pathway to automating the phenological assessment, a fundamental architectural phenomenological disconnect exists in the current approach. The state-of-the-art object detectors, from the YOLO family to the entire DETR lineage including RT-DETR, are fundamentally designed for discrete classification task \cite{carionEndtoEndObjectDetection2020a, wangYOLOv9LearningWhat2024}. These models predict a singular distinct class from a predefined class label from a predefined set (e.g., "tomato," "apple," "unripe," "ripe"). However, as established in agricultural phenology literature, maturity is not a discrete categorical state but a continuous developmental spectrum \cite{meierBBCHSystemCoding2009}. Forcing this continuous biological process into arbitrary, human-defined classes represents a fundamentally flawed approximation that discard critical information. This categorical approach not only inherits the subjectivity of the original manual labels but also loses invaluable information about the subtle, transitional states of development. An object that is 90\% ripe is treated identically to those of 60\% ripeness, a critical distinction for optimizing crop management, harvest timing and quality. This limitation reveals a crucial gap: the very architecture of modern object detectors is structurally misaligned with the nature of the problem.


To address this fundamental limitation, we propose a novel methodological framework that leverages the prediction architecture of a state-of-the-art detector to reason directly about continuous attributes rather than discrete categories . In this work, we present a significant modification to the RT-DETR architecture\cite{lvRTDETRv2ImprovedBaseline2024} by replacing its standard classification head to do binary classification fruit or not fruit and introducing a probabilistic head for maturity. This innovation enables our model to predict the maturity of produce not as a fixed class, but as a continuous value with an associated uncertainty. By doing so, we move beyond subjective, categorical labels and toward a more objective, quantitative, and informative representation of object state. Our primary contributions are:

\begin{itemize}
\item The design and implementation of a novel probabilistic maturity head for the RT-DETR architecture, enabling end-to-end training for continuous attribute estimation alongside object detection.

\item A comprehensive demonstration that predicting maturity as a continuous distribution provides a richer, more robust signal for agricultural automation tasks compared to traditional classification-based approaches.

\item Extensive experimental validation on a challenging real-world dataset, showcasing our model's superior performance in accurately assessing fruit maturity.

\end{itemize}

\section{Related Work}
Modern deep learning approaches have transformed fruit and vegetable maturity assessment across multiple crops and architectural paradigms. CNN-based methods established early foundations through techniques like feature concatenation and transfer learning. For instance, Garillos-Manliguez and Chiang \cite{s21041288} achieved F1 scores up to 0.90 for papaya maturity using multimodal RGB and hyperspectral imaging (400-900 nm) with feature concatenation across AlexNet, VGG16, and ResNet50 (though notably, this study did not use transfer learning). In contrast, Behera et al. \cite{BEHERA2021244} did use transfer learning, reporting 100\% accuracy for papaya classification with VGG19. Fine-grained banana ripeness classification across seven stages was pioneered by Zhang et al. \cite{Zhang2018} using triplet loss, and Gao et al. \cite{GAO202031} demonstrated 98.6\% accuracy for strawberry ripeness using a pretrained AlexNet CNN trained on spatial features from hyperspectral images (specifically, 530 nm grayscale images and the first three principal components).

YOLO architectures dominate real-time detection applications. For tomatoes, Liu et al. \cite{Liu2020YOLO} introduced YOLO-Tomato with circular bounding boxes achieving 96.4\% AP at 54 ms, while recent lightweight variants like Zeng et al.'s \cite{Zeng2023Lightweight} pruned YOLOv5 reduced parameters by 78\% while maintaining accuracy for edge deployment. The latest generation includes Zhao et al.'s \cite{Zhao2025YOLODGS} YOLO-DGS based on YOLOv10, achieving 80.1\% mAP50 with 26.3\% fewer parameters through D-2Detect (removal of redundant detection layers) and C2f-GB modules. Strawberry detection has similarly progressed: Chai et al. \cite{Chai2023RealTime} achieved 89\% mAP with YOLOv7 at 18 ms inference time, Pang et al. \cite{Pang2023MSYOLOv5} reached 95.6\% mAP at 76 FPS with MS-YOLOv5, and Yang et al. \cite{YANG2023108360} integrated LW-Swin Transformer with YOLOv8s, achieving 94.4\% mAP@0.5.

Vision Transformers and end-to-end detectors represent recent advances. Xiao et al. \cite{Xiao2023Fruit} demonstrated that Swin Transformer's hierarchical self-attention outperforms standard ViT for small fruit detection, while Wu et al. \cite{Wu2025ImprovedRTDETR} first applied RT-DETR to fruit ripeness with Rep Block and Efficient Multi-Scale Attention, achieving superior accuracy for precision tasks. Liu et al. \cite{agriculture15010075} optimized Deformable DETR for occluded green apple detection, achieving a 2.6\% AP improvement over the baseline model (54.1\% vs. 51.5\%) through a deformable attention mechanism and a ResNeXt backbone. However, the common detection frameworks compared in this field, such as the YOLO series and DETR, typically employ discrete classification and deterministic regression heads, which do not inherently model continuous distributions or quantify prediction uncertainty.

\subsection{Uncertainty Quantification in Computer Vision}
Reliable uncertainty estimation is crucial for safety-critical vision systems. Bayesian foundations were established by Blundell et al. \cite{Blundell2015Weight}, who introduced variational inference over network weights (Bayes by Backprop), Gal and Ghahramani \cite{pmlr-v48-gal16}, who demonstrated that dropout training corresponds to approximate Bayesian inference, enabling Monte Carlo dropout for epistemic uncertainty without architectural modifications. 

A critical conceptual advance came from Kendall and Gal \cite{Kendall2017Uncertainties}, who formalized the distinction between epistemic uncertainty (model uncertainty, reducible with more data) and aleatoric uncertainty (observation noise, irreducible). Their framework models both simultaneously through heteroscedastic outputs (predicting mean and variance) combined with Bayesian weight distributions, achieving state-of-the-art results on semantic segmentation and depth regression while improving robustness to noisy annotations.

Deep ensembles provide a simpler alternative: Lakshminarayanan et al. \cite{lakshminarayanan2017simplescalablepredictiveuncertainty} showed that training multiple networks from different initializations with proper scoring rules achieves superior calibration compared to approximate Bayesian methods while scaling to ImageNet. However, ensembles require K forward passes for K models, limiting real-time applicability.

Evidential deep learning offers computational efficiency through single-pass uncertainty estimation. Sensoy et al. \cite{NEURIPS2018_a981f2b7} introduced evidential classification using Dirichlet distributions over class probabilities, achieving unprecedented out-of-distribution detection. Amini et al. \cite{NEURIPS2020_aab08546} extended this to regression through Deep Evidential Regression, placing Normal-Inverse-Gamma priors over Gaussian likelihoods to simultaneously learn epistemic and aleatoric uncertainty for continuous outputs with an evidential regularizer that penalizes misaligned evidence.

\subsection{Probabilistic Prediction Heads for Continuous Attributes}
Beyond discrete classification, many vision tasks benefit from probabilistic continuous prediction. Mixture Density Networks \cite{Bishop1994Mixture} combine neural networks with Gaussian mixtures to represent multimodal conditional distributions, excelling at inverse problems but requiring significant parameterization.

For bounded regression targets, Beta and Dirichlet distributions provide natural parameterizations. Sadowski et al. \cite{sadowski2019neural} demonstrated neural network regression with Beta outputs for 1D simplices (suitable for [0,1] intervals like maturity percentages) and Dirichlet outputs for multi-dimensional compositional data, training via negative log-likelihood to quantify uncertainty when targets are inherently probabilistic.

Heteroscedastic regression, where networks predict input-dependent variance, has become widespread. Kendall and Gal \cite{Kendall2017Uncertainties} demonstrated Gaussian outputs for depth regression using loss $L = (1/2\sigma^2)\|y - \hat{y}\|^2 + (1/2)\log(\sigma^2)$. Recent improvements include Seitzer et al.'s \cite{seitzer2022pitfallsheteroscedasticuncertaintyestimation} $\beta$-NLL loss for stable training and Stirn et al.'s \cite{pmlr-v206-stirn23a} faithful heteroscedastic regression ensuring proper calibration.

Label distribution learning for ordinal attributes has proven valuable for age estimation. Gao et al. \cite{ijcai2018p99} introduced Deep Label Distribution Learning (DLDL), which represents ages as Gaussian distributions; however, in this original framework, the distribution's variance ($\sigma$) was a fixed hyperparameter rather than a learned variable, while Niu et al. \cite{Niu_2016_CVPR} employed ordinal regression through K-1 binary outputs for K ranks. or gaze estimation, Kellnhofer et al. \cite{kellnhofer2019gaze360physicallyunconstrainedgaze} used quantile regression for 3D uncertainty, and Prokudin et al. \cite{Prokudin2018Deep} applied von Mises distributions for angular head pose with circular statistics. These applications demonstrate the broad utility of distributional predictions across vision tasks with continuous attributes, ambiguous annotations, or inherent uncertainty.

 \section{Foundational Architecture: RT-DETR}
Our work builds upon the Real-Time Detection Transformer (RT-DETR)\cite{zhaoDETRsBeatYOLOs2024}, a state-of-the-art, end-to-end object detector. The RT-DETR model represents a significant leap forward by successfully extending the DETR \cite{carionEndtoEndObjectDetection2020} architecture to real-time applications, thereby challenging the long-standing dominance of YOLO-based models \cite{redmonYouOnlyLook2016,wangYOLOv7TrainableBagofFreebies2023} . To establish a clear foundation for the novel contributions presented in this paper, this section provides a comprehensive overview of the core architectural components of the RT-DETR framework.

The RT-DETR framework is composed of three primary components: a CNN backbone for feature extraction, an efficient hybrid encoder for processing multi-scale features, and a Transformer decoder for object query refinement and prediction.

\subsection{Backbone and Multi-Scale Feature Extraction}
The RT-DETR model utilizes a standard ResNet series backbone \cite{heDeepResidualLearning2015} to extract a rich hierarchy of feature maps from an input image. In line with modern detector designs, it leverages multi-scale features to effectively handle objects of varying sizes. Specifically, the framework processes features from the last three stages of the backbone, denoted as S3, S4, and S5. This multi-scale strategy is not only crucial for scale-invariant object detection but has also been demonstrated to accelerate training convergence and improve overall performance in DETR-based models\cite{carionEndtoEndObjectDetection2020a,zhuDeformableDETRDeformable2021}.

\subsection{Efficient Hybrid Encoder}
A pivotal innovation within RT-DETR is its efficient hybrid encoder, which is meticulously designed to mitigate the significant computational bottleneck typically associated with Transformer encoders in object detection. The encoder architecture elegantly decouples feature processing into two specialized modules: the Attention-based Intra-scale Feature Interaction (AIFI) and the CNN-based Cross-scale Feature Fusion (CCFF).

The AIFI module strategically applies a self-attention mechanism exclusively to the highest-level feature map (S5). This targeted application of attention allows the model to capture complex semantic relationships and global context among potential object instances with high efficiency. Conversely, the CCFF module employs a lightweight, PANet-style\cite{liuPathAggregationNetwork2018} structure to efficiently fuse features across different scales (S3, S4, and the output of AIFI). By decoupling the computationally intensive intra-scale self-attention from the more efficient CNN-based cross-scale fusion, the hybrid encoder achieves a substantial reduction in latency while maintaining feature representation quality.

\subsection{Query Selection and Transformer Decoder}
To produce the final predictions, RT-DETR employs a sophisticated query selection and decoding process that obviates the need for hand-crafted components like anchors or NMS post-processing.

\textbf{Uncertainty-Minimal Query Selection}: Moving away from the learnable object queries of the original DETR, RT-DETR initializes its decoder queries directly from the encoder's output features. It employs an uncertainty-minimal query selection scheme, which selects the top-K encoder features that exhibit the highest consistency—and thus, least uncertainty—between their predicted class and location. This data-driven approach provides the decoder with high-quality initial object proposals, simplifying the optimization process.

\textbf{Iterative Refinement Decoder}: The selected queries are then fed into a standard Transformer decoder. Equipped with auxiliary prediction heads at each layer, the decoder iteratively refines the object queries through cross-attention with the encoder's output features. This iterative refinement allows the model to fine-tune both the class and bounding box coordinates for each object. The final output is a one-to-one set of predictions, which elegantly circumvents the need for post-processing steps like Non-Maximum Suppression (NMS), a common bottleneck in other real-time detectors.

\section{Methodology}

The proposed methodology is built upon the RT-DETR v2 architecture, a state-of-the-art, real-time end-to-end object detector. The foundational framework, which consists of a backbone for feature extraction, a hybrid encoder, and a query-based decoder, is enhanced to perform probabilistic regression for a continuous biological trait i.e. maturity in addition to its core single-class "objectness" detection task.

The core methodological approach transforms a deterministic model into a probabilistic one by predicting the parameters of a full probability distribution; is inspired by the framework for t-Distributed Neural Networks (TDistNNs) proposed by \cite{pourkamali-anarakiProbabilisticNeuralNetworks2025}. However, this work diverges in the choice of the statistical distribution to better suit the specific nature of the target variable. Whereas the t-distribution is ideal for modeling unbounded variables with potential outliers, the biological trait of 'maturity' is a continuous variable naturally constrained to a bounded interval of [0,1]. Therefore, the Beta distribution was selected as a more statistically appropriate model for this task.

This is achieved through three primary modifications: (1) the integration of a dedicated probabilistic prediction head to output the parameters of a Beta distribution, (2) a novel label assignment strategy that incorporates maturity cost, and (3) a tailored loss function designed to quantify prediction uncertainty.

\subsection{Model Architecture Modifications}
To enable the model to predict maturity not as a fixed value but as a probability distribution, the architecture is extended with a dedicated prediction head.

\subsubsection{Beta Distribution for Bounded Traits}
Maturity is a continuous variable normalized to a bounded interval [0,1], where 0 represents a fully immature state and 1 represents a fully ripe state. The Beta distribution is uniquely suited for modeling such bounded quantities. Its probability density function is defined by two positive shape parameters, $\alpha$ and $\beta$, which control the distribution's form:

$$p(y|\alpha, \beta) = \frac{y^{\alpha-1}(1-y)^{\beta-1}}{B(\alpha, \beta)}$$

where y is the maturity value and B($\alpha, \beta$) is the Beta function. By predicting $\alpha$ and $\beta$, the model can represent its uncertainty about an object's maturity. A high-variance distribution (low $\alpha$ and $\beta$) indicates low confidence, while a low-variance distribution indicates high confidence.

\subsubsection{Probabilistic Prediction Head}

A new prediction head, termed the Probabilistic Maturity Head, is integrated into the model. This head is a multi-layer perceptron (MLP) that takes the hidden state from the transformer's feature representation as input. For each object query, it outputs two scalar values corresponding to the parameters of a Beta distribution. To ensure that $\alpha$ and $\beta$ are always positive and greater than 0.5 (for numerical stability), the raw network outputs undergo a softplus transformation followed by an offset:

$$\alpha = \text{softplus}(\hat{y}_\alpha) + 0.5$$

$$\beta = \text{softplus}(\hat{y}_\beta) + 0.5$$

This head is appended to the encoder and each layer of the transformer decoder, enabling deep supervision across all stages of the detection process.

\subsection{Loss function}
The training objective is composed of losses for object localization, objectness classification, and the newly introduced probabilistic maturity regression.

The Probabilistic Maturity Head is trained by maximizing the Log-Likelihood of the ground-truth maturity value under the predicted Beta distribution. First, discrete ground-truth labels (e.g., "immature," "half-ripe," "ripe") are mapped to continuous target values $y_{\text{target}} \in [0, 1]$. The maturity loss is then formulated as:

$$
\mathcal{L}_{\text{maturity}} = - \log(p(y_{\text{target}}|\alpha, \beta)) + \lambda_{\text{reg}}(\alpha + \beta)
$$

This loss encourages the model to produce distributions where the ground-truth value has a high probability. A small regularization term, weighted by $\lambda_{\text{reg}}$, is added to prevent the distribution parameters from growing excessively large, which can lead to overconfident and unstable predictions.

\subsubsection{Label Assignment with Maturity Cost}
The label assignment process, which matches predictions to ground-truth objects, is crucial for training. The standard Hungarian algorithm is employed, but the cost matrix is modified to incorporate the maturity prediction quality. The total cost C for matching a prediction i to a ground-truth object j is a weighted sum of classification, bounding box, and maturity costs:

$$
\mathcal{C}(i, j) = \lambda_{\text{cls}}\mathcal{C}_{\text{cls}}(i, j) + \lambda_{\text{L1}}\mathcal{C}_{\text{L1}}(i, j) + \lambda_{\text{giou}}\mathcal{C}_{\text{giou}}(i, j) + \lambda_{\text{mat}}\mathcal{C}_{\text{mat}}(i, j)
$$

Here $C_{\text{cls}}$, $C_{\text{L1}}$, and $C_{\text{giou}}$ are the standard focal loss, L1 distance, and Generalized IoU costs respectively. The novel term, $C_{\text{mat}}$ is the NLL of the maturity prediction. This ensures that the matching process favors predictions that are accurate in both physical location and probabilistic maturity estimation.

\subsubsection{Final Loss Composition}
The final training objective is a weighted sum of all losses computed on the matched pairs from the Hungarian assignment. The objectness classification is optimized using Varifocal Loss (VFL), while localization is handled by L1 and GIoU losses. The complete loss function is:

$$ \mathcal{L} = \lambda_{\text{vfl}}\mathcal{L}_{\text{vfl}} + \lambda_{\text{bbox}}\mathcal{L}_{\text{bbox}} + \lambda_{\text{giou}}\mathcal{L}_{\text{giou}} + \lambda_{\text{maturity}}\mathcal{L}_{\text{maturity}}
$$

This composite loss is applied to the outputs of the final decoder layer as well as the auxiliary outputs from all intermediate decoder layers and the encoder, ensuring robust feature learning throughout the network.

\section{Experiment and Results}

\subsection{Dataset}

To rigorously evaluate our proposed method, we utilize the Multimodal image dataset of tomato fruits with different maturity \cite{zhangyuMultimodalImageDataset2023}, a large-scale public benchmark designed to simulate real-world agricultural scenarios. This dataset is particularly well-suited for our task as it presents several significant challenges inherent to robotic harvesting.

Key Dataset Characteristics: The benchmark contains 4,000 sets of images, capturing tomato clusters with dense distribution and asynchronous maturity. Crucially, it provides images under four highly varied and challenging illumination conditions: (1) natural light, (2) artificial light, (3) weak light, and (4) sodium yellow light. This photometric variance is a primary obstacle in practical deployment, as it can severely degrade the quality of visual features and confound traditional models.

Annotations and Task: All images are meticulously annotated for both target detection and semantic segmentation, with fruits classified into three key agronomical stages: unripe, half-ripe, and ripe. This multi-class labeling is essential for our objective of maturity-aware harvesting.

Experimental Protocol: While the full dataset provides multimodal data (RGB, Depth, and Near-Infrared), we deliberately constrain our model to use only the RGB modality. This design choice is motivated by the practical goal of developing a cost-effective and generalizable system that does not depend on expensive or specialized active-sensing equipment. By demonstrating high performance on this challenging, RGB-only subset, we validate our model's robustness and its ability to extract discriminative features for maturity classification despite significant noise, complex occlusions, and severe illumination shifts.

\begin{figure}[t]
  \centering
  \includegraphics[width=0.85\linewidth]{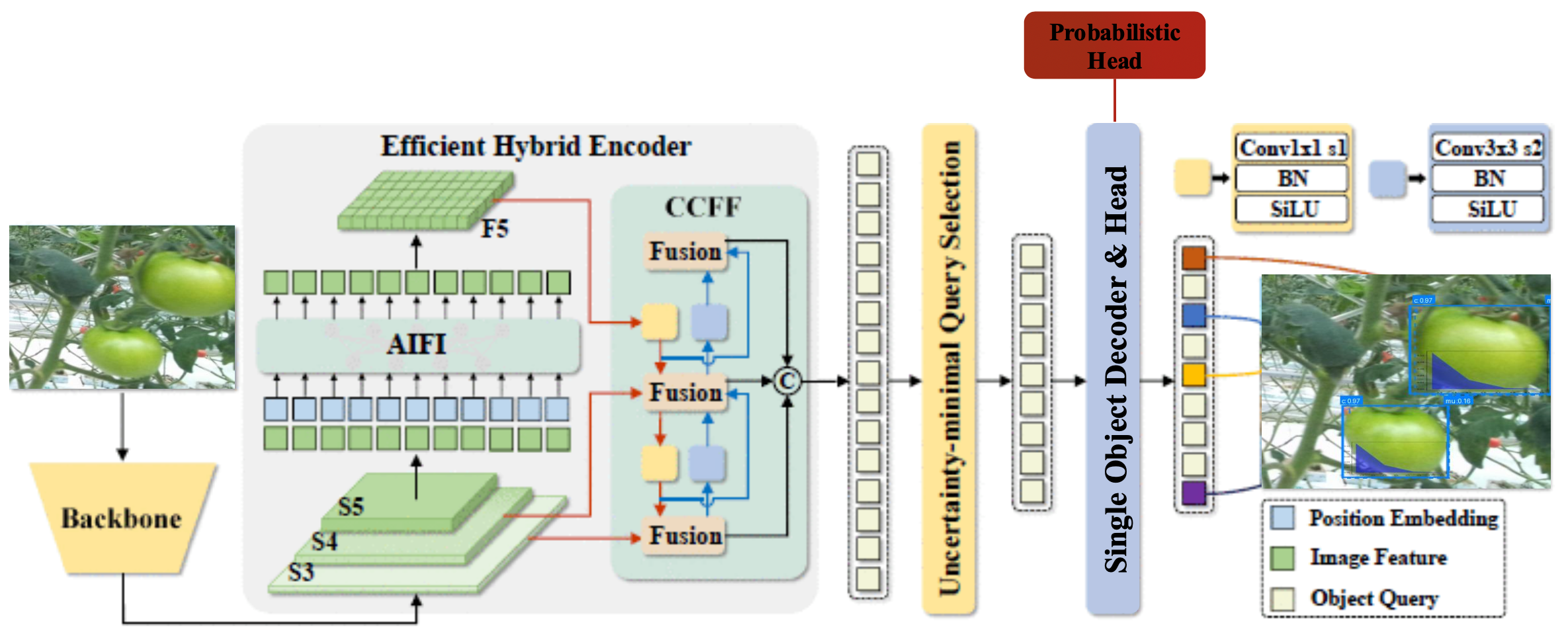}
  \caption{Architecture of the probabilistic maturity head integrated into RT-DETRv2. The head predicts Beta distribution parameters ($\alpha,\beta$) per query and is applied to encoder and decoder layers for deep supervision.}
  \label{fig:probabilistic_head_arch}
\end{figure}

 \begin{table}[ht]
  \caption{Comparison with state-of-the-art methods on the Multimodal Tomato Dataset.}
  \label{tab:state_of_the_art_comparison}
  \centering
  \begin{tabular}{@{}llccccccc@{}}
    \toprule
    Method & Backbone &   $AP_{50:95}$ & $AP_{50}$ & $AP_{75}$ & $AP_{S}$ & $AP_{M}$ & $AP_{L}$ \\
    \midrule
    Anchor-DETR \cite{wangAnchorDETRQuery2022}   & ResNet-50 & 76.1 & 96.2 & 89.3 & 44.3 & 78.3 & 89.7 \\
    Yolo v7 \cite{wangYOLOv7TrainableBagofFreebies2023}       & -         & 79.5 & 96.9 & -    & -    & -    & -    \\
    DAB-Def-DETR \cite{liuDABDETRDynamicAnchor2022}   & ResNet-50 & 79.5 & 95.3 & 90.6 & 47.5 & 82.6 & 93.3 \\
    MS-DETR \cite{zhaoMSDETREfficientDETR2024}       & ResNet-50 & 80.4 & 96.1 & 91.5 & 48.5 & 82.6 & 93.7 \\
    DINO \cite{zhangDINODETRImproved2022}          & ResNet-50 & 82.1 & 96.0 & 92.1 & 49.8 & 84.4 & 93.9 \\
    Salience-DETR \cite{houSalienceDETREnhancing2024} & ResNet-50 & 82.6 & 96.0 & 91.7 & 51.3 & 84.7 & 93.8 \\
    Relation-DETR \cite{houRelationDETRExploring2025} & ResNet-50 & 82.7 & 96.2 & 92.2 & 52.0 & 84.7 & 93.2 \\
    EG-DETR \cite{yaoEdgeGuidedDETRModel2025}            & ResNet-50 & 83.7 & 96.5 & 92.6 & 54.0 & 85.7 & 95.4 \\
    \midrule
     FruitProm (Ours) & ResNet-50      & 85.6 & 96 & 93 & 55.2 & 87.2 & 95.9 \\
    \bottomrule
  \end{tabular}
\end{table}

\subsection{Results}

Table \ref{tab:state_of_the_art_comparison} presents a comprehensive comparison of our proposed  FruitProm (Probabilistic Maturity DETR) against eight state-of-the-art object detection architectures, all evaluated on the Multimodal Tomato Dataset under identical experimental conditions. All competing methods utilize ResNet-50 backbones for fair comparison.

Evaluation Metrics: Following standard object detection protocols, we report mean Average Precision (mAP) at multiple Intersection over Union (IoU) thresholds: $AP_{50:95}$ (averaged over IoU thresholds from 0.5 to 0.95 with 0.05 increments), $AP_{50}$ (IoU threshold of 0.5), and $AP_{75}$ (IoU threshold of 0.75). Additionally, we report mAP stratified by object scale: $AP_{S}$ (small objects with area $< 32^2$), $AP_{M}$ (medium objects with $32^2 < \text{area} < 96^2$), and $AP_{L}$ (large objects with area $> 96^2$). 

\subsection{Quantitative Analysis}

Our FruitProm achieves a new state-of-the-art $AP$ of 85.6\%, outperforming the previous best method (EG-DETR) by a significant 1.9 percentage points (Table 1). This improvement is particularly noteworthy given that FruitProm jointly optimizes an auxiliary probabilistic maturity regression head—indicating that the additional, more principled learning objective enriches the model's representational capacity rather than diluting it. The model's consistency is high, with an $AP_{50}$ of 96.0\%, while the strong $AP_{75}$ of 93.0\% confirms that our model generates precisely localized and well-calibrated bounding boxes, even under strict matching criteria.

This superiority is consistent across all object scales. Most notably, FruitProm achieves an $AP_S$ of 55.2\%, a 1.2 pp improvement over the prior best, reflecting its enhanced capability in detecting small, visually ambiguous fruits. This is critical in agricultural scenarios, where early fruit detection—often characterized by smaller object sizes—is essential for predictive yield modeling. For medium and large instances, our model attains $AP_M$ of 87.2\% and $AP_L$ of 95.9\%, respectively, both exceeding existing state-of-the-art results.

Furthermore, when compared to architecturally similar DETR variants—such as Anchor-DETR, DAB-Def-DETR, MS-DETR, DINO, Salience-DETR, and Relation-DETR—FruitProm consistently delivers performance gains ranging from 3.0 to 9.5 pp in $AP$. This suggests that our probabilistic formulation contributes benefits that are orthogonal to recent architectural refinements in query initialization or attention modulation. Interestingly, FruitProm also surpasses EG-DETR, which explicitly integrates edge-guided priors for boundary refinement. This finding indicates that the uncertainty-aware probabilistic head in FruitProm implicitly learns geometric consistency and boundary confidence without the need for handcrafted edge constraints, acting as a powerful regularization mechanism.


\begin{figure}[t]
  \centering
  \includegraphics[width=0.85\linewidth]{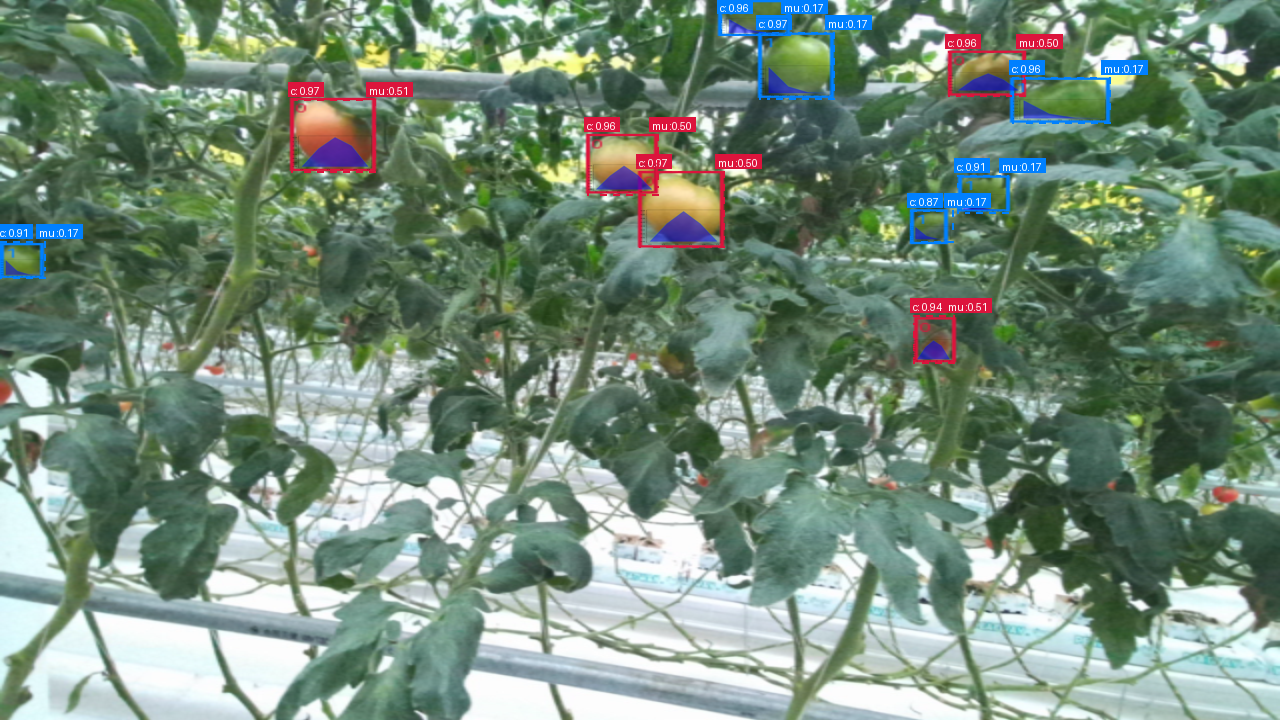}
  \caption{Qualitative results of FruitProm on tomato image. The predicted maturity distributions illustrate the model's uncertainty handling across different ripening stages.}
  \label{img:qualitative_results}
\end{figure}

\subsection{Qualitative Analysis}

Beyond quantitative metrics, the practical utility of our model is illustrated through qualitative examples (Fig. \ref{img:qualitative_results}). Figure \ref{img:qualitative_results} showcases our model's predictions for tomatoes at various ripening stages. For a fruit transitioning from unripe to ripe, our model correctly predicted a wider, flatter probability distribution, capturing the inherent biological ambiguity. In contrast, a classification-based model is forced to make a discrete, and often incorrect, choice. This highlights the superior representational power of our continuous approach.

We also demonstrate how the learned uncertainty correlates with challenging scenarios. For instance, heavily occluded or poorly illuminated fruits yielded predictions with higher variance (a wider distribution). This uncertainty measure is a critical output for downstream robotic applications, enabling a system to, for example, deprioritize harvesting fruits for which it has low confidence, thereby minimizing errors.

\section{Discussion}


The results collectively demonstrate that probabilistic reasoning and maturity modeling can substantially enhance object detection performance—not merely as an auxiliary task, but as a structural prior that enriches the model's spatial and semantic understanding.

From a theoretical standpoint, the probabilistic head acts as an inductive bias, encouraging the model to reason over prediction distributions rather than point estimates. This yields more calibrated confidence maps, particularly beneficial in agricultural domains where occlusion, lighting variation, and visual ambiguity are frequent.

Practically, FruitProm's joint estimation of object boundaries and maturity states opens a promising avenue toward multifunctional perception models capable of supporting both visual phenotyping and agronomic decision-making. Its performance improvements—especially on small and medium objects—demonstrate the feasibility of end-to-end, uncertainty-aware perception pipelines for precision agriculture.

\section{Conclusion}
In this work, we challenged the conventional classification-based approach to maturity estimation and proposed a more principled, continuous, and probabilistic framework. Our model, FruitProm, sets a new state-of-the-art on the Multimodal Tomato Dataset, demonstrating that a more biologically plausible problem formulation leads to superior performance. The ability to predict not just a maturity state but also the uncertainty associated with it paves the way for more robust and intelligent automated systems in agriculture.

Future work could further explore integrating temporal cues (for fruit growth tracking) or spectral modalities (e.g., hyperspectral maturity sensing) into the probabilistic framework, thereby bridging the gap between semantic detection and physiological interpretation.

\printbibliography

\end{document}